\documentclass[sigconf, noacm]{acmart}
\AtBeginDocument{%
  }

\usepackage{xcolor}

\newcommand{\ds}[1]{{\color{purple}#1}}
\usepackage{soul} 

\usepackage{multirow}
\usepackage{makecell} 
\usepackage{booktabs}
\usepackage{listings}
\usepackage[T1]{fontenc}
\usepackage{tcolorbox}
\usepackage{paralist}

\lstdefinestyle{logstyle}{
  basicstyle=\ttfamily\footnotesize,
  breaklines=true,
  columns=fullflexible,
  keepspaces=true,
  showstringspaces=false,
  frame=single,
  rulecolor=\color{black!25},
  backgroundcolor=\color{black!3},
  moredelim={**[is][\color{red!70!black}]{<<R>>}{<<END>>}},
  moredelim={**[is][\color{green!50!black}]{<<G>>}{<<END>>}},
}

\definecolor{eclipseStrings}{RGB}{42,0,255}
\definecolor{eclipseKeywords}{RGB}{127,0,85}
\definecolor{jsonKeys}{RGB}{165, 42, 42} 
\definecolor{jsonStrings}{RGB}{0, 128, 0} 
\definecolor{jsonNumbers}{RGB}{255, 140, 0} 

\lstdefinelanguage{json}{
    basicstyle=\ttfamily\footnotesize,
    numbers=left,
    numberstyle=\tiny\color{gray},
    stepnumber=1,
    numbersep=4pt,
    tabsize=1,
    showstringspaces=false,
    breaklines=true,
    frame=lines,
    backgroundcolor=\color{gray!5}, 
    stringstyle=\color{jsonStrings},
    literate=
     *{0}{{{\color{jsonNumbers}0}}}{1}
      {1}{{{\color{jsonNumbers}1}}}{1}
      {2}{{{\color{jsonNumbers}2}}}{1}
      {3}{{{\color{jsonNumbers}3}}}{1}
      {4}{{{\color{jsonNumbers}4}}}{1}
      {5}{{{\color{jsonNumbers}5}}}{1}
      {6}{{{\color{jsonNumbers}6}}}{1}
      {7}{{{\color{jsonNumbers}7}}}{1}
      {8}{{{\color{jsonNumbers}8}}}{1}
      {9}{{{\color{jsonNumbers}9}}}{1}
      {:}{{{\color{eclipseKeywords}{:}}}}{1}
      {,}{{{\color{eclipseKeywords}{,}}}}{1}
      {\{}{{{\color{eclipseKeywords}{\{}}}}{1}
      {\}}{{{\color{eclipseKeywords}{\}}}}}{1}
      {[}{{{\color{eclipseKeywords}{[}}}}{1}
      {]}{{{\color{eclipseKeywords}{]}}}}{1},
}

\definecolor{darkgreen}{RGB}{135, 169, 107}
\definecolor{lightgreen}{RGB}{245, 255, 250}
\definecolor{darkpink}{HTML}{D88694}
\definecolor{lightpink}{HTML}{F7E7E9}

\usepackage{tabularx}
\usepackage{graphicx}

\usepackage{xurl}

\newcommand{\model}[1]{\begingroup\urlstyle{tt}\nolinkurl{#1}\endgroup}

\begin{document}

\title{{\sc \textbf{ReplicatorBench}}: Benchmarking LLM Agents for Replicability in Social and Behavioral Sciences}


\author{Bang Nguyen}
\authornote{The authors contributed equally to this research.}
\affiliation{%
  \institution{University of Notre Dame}
  \department{Computer Science and Engineering}
  \city{Notre Dame}
  \state{IN}
  \country{United States}
}
\email{bnguyen5@nd.edu}

\author{Dominik Soós}
\authornotemark[1]
\affiliation{%
  \institution{Old Dominion University}
  \department{Computer Science}
  \city{Norfolk}
  \state{VA}
  \country{United States}
}
\email{dsoos001@odu.edu}

\author{Qian Ma}
\affiliation{%
  \institution{Pennsylvania State University}
  \department{Information Sciences and Technology}
  \city{University Park}
  \state{PA}
  \country{United States}
}
\email{qfm5033@psu.edu}

\author{Rochana R. Obadage}
\affiliation{%
 \institution{Old Dominion University}
 \department{Computer Science}
 \city{Norfolk}
  \state{VA}
  \country{United States}
}
\email{oruma001@odu.edu}

\author{Zack Ranjan}
\affiliation{%
  \institution{Pennsylvania State University}
  \department{Information Sciences and Technology}
  \city{University Park}
  \state{PA}
  \country{United States}
}
\email{sjr6223@psu.edu}

\author{Sai Koneru}
\affiliation{%
  \institution{Pennsylvania State University}
  \department{Information Sciences and Technology}
  \city{University Park}
  \state{PA}
  \country{United States}
}
\email{sdk96@psu.edu}



\author{Timothy M. Errington}
\affiliation{%
  \institution{Center for Open Science}
  \city{Washington}
  \state{DC}
  \country{United States}
}
\email{tim@cos.io}

\author{Shakhlo Nematova}
\affiliation{%
  \institution{Center for Open Science}
  \city{Washington}
  \state{DC}
  \country{United States}
}
\email{shakhlo@cos.io}

\author{Sarah Rajtmajer}
\affiliation{%
  \institution{Pennsylvania State University}
  \department{Information Sciences and Technology}
  \city{University Park}
  \state{PA}
  \country{United States}
}
\email{smr48@psu.edu}

\author{Jian Wu}
\affiliation{%
  \institution{Old Dominion University}
  \department{Computer Science}
  \city{Norfolk}
  \state{VA}
  \country{United States}
}
\email{jwu@cs.odu.edu}

\author{Meng Jiang}
\affiliation{%
 \institution{University of Notre Dame}
  \department{Computer Science and Engineering}
 \city{Notre Dame}
 \state{IN}
 \country{United States}
}
\email{mjiang2@nd.edu}

\authorsaddresses{}
\renewcommand{\shortauthors}{Bang Nguyen et al.}

\begin{abstract}
The literature has witnessed an emerging interest in developing and evaluating AI agents for automated assessment of research claims in scientific papers.
Existing benchmarks focus primarily on the computational aspect of this task, testing agents' ability to reproduce or replicate research outcomes when having access to the code and data.
This setting, while foundational, (1) fails to capture the inconsistent availability of new data for replication as opposed to reproduction, 
and (2) lacks ground-truth diversity by focusing exclusively on fully reproducible or replicable papers, thereby failing to evaluate an agent's ability to identify non-replicable research.
Furthermore, most benchmarks only evaluate the final reproducibility or replicability outcomes without an evaluation of the process. 
In response, we introduce {\sc ReplicatorBench}, an end-to-end benchmark, including human-verified replicable and non-replicable research claims in social and behavioral sciences, for evaluating AI agents in research replication across three stages: (1) extraction of relevant information and retrieval of replication data; (2) design and execution of computational experiments; and (3) interpretation of replication results, allowing a test of AI agents' capability to mimic the activities of human replicators in real world.
To set a baseline of AI agents' capability, we develop {\sc ReplicatorAgent}, an agentic framework equipped with necessary tools like web search and iterative interaction with sandboxed environments, to accomplish tasks in {\sc ReplicatorBench}.
We evaluate {\sc ReplicatorAgent} across four underlying large language models (LLMs), as well as different design choices of programming language and levels of code access.
Our findings reveal that while current LLM agents are capable of effectively designing and executing computational experiments, they struggle with retrieving new data, necessary to replicate a claim.
All code and data are publicly available at: \href{https://github.com/CenterForOpenScience/llm-benchmarking}{https://github.com/CenterForOpenScience/llm-benchmarking}.
\end{abstract}


\setcopyright{none}
\settopmatter{printacmref=false}
\settopmatter{printfolios=false}
\acmDOI{}
\acmISBN{}
\acmConference{}{}{}
\acmBooktitle{}
\acmYear{}
\copyrightyear{}
\acmPrice{}


\maketitle

\begingroup\small\noindent\raggedright\textbf{Resource Availability:}\\
The prompts, structured output schemas, and supplementary materials associated with this paper are publicly available in an archival Zenodo artifact at \url{https://doi.org/10.5281/zenodo.20506946}.
\endgroup

\begin{figure*}[ht!]
    \centering
    \includegraphics[width=0.9\textwidth]{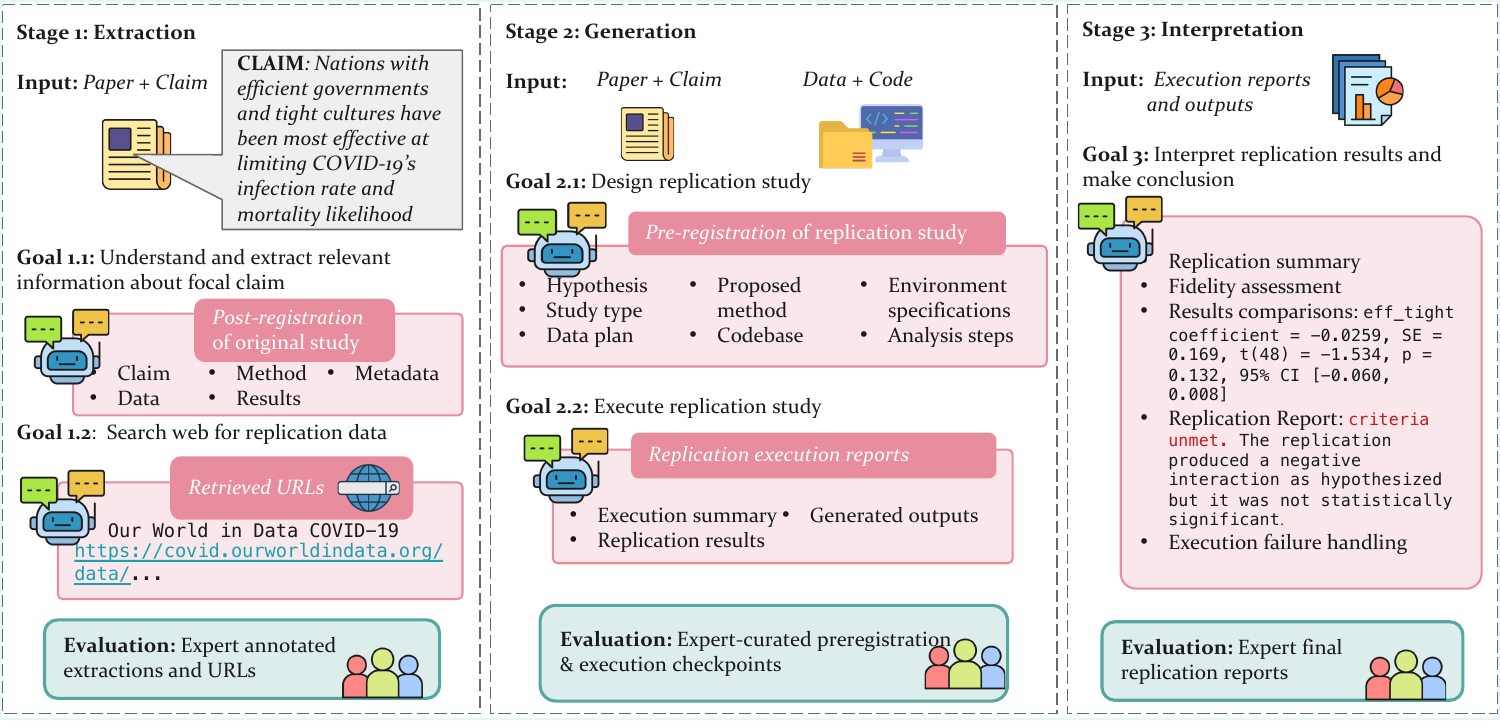}
    \vspace{-0.15in}
    \caption{Given a research paper and a focal claim, {\sc \textbf{ReplicatorBench}} decomposes the replication process into three stages.
    (1) \textit{Extraction} assesses the agent's ability to gather relevant information about the claim and retrieve data resources for replication;
    (2) \textit{Generation} assesses the computational capacity of an agent to generate and execute code for replication;
    (3) \textit{Interpretation} assesses the agent's ability to interpret computational outputs and make conclusions regarding the focal claim's replicability.
    }
    \label{fig:task-design}
    \vspace{-0.14in}
    \Description[Overview of each stage of our work.]{Given a research paper and a focal claim, {\sc \textbf{ReplicatorBench}} decomposes the replication process into three stages.}
\end{figure*}
\section{Introduction}

Recent advances in large language models (LLMs) have enabled autonomous agents that can read documents, write and execute code, use tools, and operate in sandboxed environments ~\cite{yao2022react,wang2024executable}.
These agents have been adopted into various stages of the research lifecycle, from idea creation, experimental design, and execution to results synthesis and paper writing~\cite{luo2025llm4sr,ren2025towards,majumder2024discoverybench}.
In social sciences, there has been an increasing interest in the development and testing of LLM agents for research reproduction and replication \cite{siegel2024core,hu2025repro}.
While research \textit{reproduction} attempts to confirm whether the same findings can be obtained using the original data samples and methods, \textit{replication} investigates the generalizability of methods when applied to new data samples~\cite{engineering2019reproducibility,nosek2020replication}.
These tasks, while essential to academic and research integrity, are resource-intensive, providing a promising use case for autonomous LLM agents.

Several benchmarks have been proposed to develop and evaluate LLM agents on research reproduction tasks, including CORE-Bench \cite{siegel2024core}, PaperBench \cite{starace2025paperbench}, and REPRO-Bench~\cite{hu2025repro}.
LLM agents have the potential to assess the reproducibility or replicability of research claims in a controlled setting.
By definition, replication requires researchers to locate or construct a new data sample \cite{nosek2020replication}.
Existing benchmarks operate under the assumption that the new data sample is readily available to the agent \cite{siegel2024core,starace2025paperbench,hu2025repro}.
There are no benchmarks designed to assess agents' capability to replicate research claims in a setting where a new data sample must be retrieved in advance.
To avoid error propagation in later stages, the new-sample data are curated and packaged for the agent rather than collected live from the open web. In addition, most existing benchmarks focus on evaluating the final reproducibility or replicability outcome without considering the human replicators' process. 

To fill these gaps, we introduce {\sc ReplicatorBench}, a benchmark for replicating published research claims in social and behavioral sciences (SBS). The benchmark includes 39 instances, each corresponding to a paper with a human expert replication report produced in the SCORE project \cite{alipourfard2021systematizing}.
SCORE (Systemizing Confidence in Open Research and Evidence) is a DARPA program, in partnership with the Center for Open Science, that coordinated a large-scale extraction of social-behavioral claims and human reproduction and replication efforts to assess each claim's credibility \cite{alipourfard2021systematizing}.

In SCORE, replication studies are documented as an end-to-end process rather than a single report, starting from preregistering the replication plan with criteria for replicability, Internet-retrieved data sources, and proposed methodology, to conducting the replication study with documented code and data, and finally creating a report to assess the claim's replicability. 
This process-oriented documentation allows evaluation of the agent's behavior at each stage of a replication study with a hierarchy of checkpoints rather than only the end outcome \cite{willroth2024best}. Here, a checkpoint is a point during the replication process at which a result is produced and evaluated.

The agent replication outcome of each instance in {\sc ReplicatorBench} falls into one of two categories: (1) \textit{criteria met}, where the preregistered criteria are satisfied on the replication sample, or (2) \textit{criteria unmet} otherwise.
This binary setting is aligned with human replication results and adopted by existing replicability benchmarks. To evaluate agent performance, we measure the alignment between the agent's concluded outcome and the ground-truth human outcome provided by the SCORE project.
{\sc ReplicatorBench} supports fine-granular evaluation and gives AI agents partial credits through evaluating $3,128$ gradable checkpoints across three main stages of 39 instances: {Extraction}, {Generation}, and {Interpretation} (illustrated in Figure~\ref{fig:task-design} and detailed in $\S$\ref{sec:task-design}).

As a baseline study, we develop {\sc ReplicatorAgent}, a tool-using replication agentic framework capable of inspecting directories, reading and editing files with minimal diffs, resolving dependencies, and rerunning analyses until it produces interpretable outputs or reaches a well-justified conclusion (criteria met or unmet).
We also study a practical trade-off for replication agents: executing native non-Python codes (e.g., R) provided by the original authors or human replicators versus executing Python codes translated from non-Python codes. 
We found that this practice streamlines the replication process by reducing installations of language compilers and software dependencies, which resulted in fewer execution failures. 
The trade-off is that it may introduce information loss that produces incorrect outputs, which in turn negatively affects the reliability of the replicability assessment. 

Our contributions are threefold.
\begin{compactenum}
    \item We introduce {\sc ReplicatorBench}, a novel benchmark consisting of 39 instances targeting evaluating AI agents' capability of replicating research claims in scientific literature in social and behavioral sciences, mimicking the real-world setting in which the new data for replication is not given. In addition to the binary outcome, the benchmark breaks the replication process into three stages, which encapsulate a total of 3,128 gradable checkpoints for partial credits.
    \item We develop {\sc ReplicatorAgent}, a tool-using agentic framework and a sandbox orchestration setup that supports iterative debugging and end-to-end replication attempts.
    \item We evaluate {\sc ReplicatorAgent} against {\sc ReplicatorBench} under several configurations, such as the choice of LLMs and whether to translate native codes to Python. 
    We highlight where today's agents succeed and fail, and the root causes of failures in the replication process, and shed light on the real-world replication challenges. 
\end{compactenum}

\section{Related Work}

\subsection{Research Reproducibility and Replicability}

Existing research has revealed significant evidence regarding the lack of replicability in fields such as psychology \cite{doi:10.1126/science.aac4716} and economics \cite{camerer2016evaluating}. Authors have explored the connections (or lack thereof) between replicability and factors such as authors' experience, prestige, and the availability of data and code \cite{brodeur2024mass}.
Recent studies have also acknowledged the time- and resource-consuming nature of manual replication and motivated the need for automated methods to replicate research claims  \cite{alipourfard2021systematizing}. 
ReplicationBench \cite{ye2025replicationbench} offers a principled benchmark to test and evaluate such methods. 

\subsection{Benchmarking LLM Agents for Assessing Reproducibility and Replicability}

Existing literature on this subject predominantly revolves around benchmarking LLMs' computational capacity for assessing reproducibility and replicability.
For example, {LMR-BENCH} \cite{yan2025lmr} withholds key implementation components from language modeling papers and evaluates whether LLMs can reconstruct them.
Additionally, LLMs have also been evaluated on their ability to solve coding problems in machine learning research in {ResearchCodeBench} \cite{hua2025researchcodebench}, in natural sciences in {SciCode} \cite{tian2024scicode}, and in natural language processing in {SciReplicate-Bench} \cite{xiang2025scireplicate}.
Benchmarks like {CoreBench} \cite{siegel2024core}, {PaperBench} \cite{starace2025paperbench}, and {ReplicationBench} \cite{ye2025replicationbench} investigate autonomous computational reproduction or replication, testing self-driven AI agents continuously from code implementation to execution and debugging without human intervention.

Replicating research claims requires mapping a focal claim to a prespecified (statistical) criterion to determine whether that criterion is met in a new dataset, using the same analytical methods. 
{CORE-Bench} \cite{siegel2024core}, a benchmark dataset consisting of 90 papers in Computer Science, Biology, and Social Sciences, addresses this requirement by introducing a subtask of interpreting execution outputs. 
However, {CORE-Bench} only includes fully reproducible claims, and therefore, cannot be used for evaluating the agents' capability for irreproducible claims. 
{REPROBench} \cite{hu2025repro} is a benchmark consisting of 112 papers in social science. 
Although papers in this benchmark include different levels of reproducibility to the test set, the benchmark situates the reproduction task within a controlled setting where agents conveniently have access to both code and data.
Our {\sc ReplicatorBench} addresses this gap by introducing a resource retrieval task as a first step to the challenge commonly faced by researchers during replication. 
Furthermore, our {\sc ReplicatorBench} differs from most existing benchmarks by evaluating the replication process, starting from information extraction from papers until interpreting computational results to draw conclusions. 
It is the first replication benchmark that provides fine-grained checkpoints to score the AI agent's capability to replicate research claims in social and behavioral sciences. 
Table~\ref{tab:benchmarks} provides a comparison between existing benchmarks and our work.

\section{{\sc \textbf{ReplicatorBench}}}

\renewcommand{\arraystretch}{1} 
\begin{table*}[ht]

\centering
\small
\begin{tabularx}{\textwidth}{lXXXX} 
\toprule
\textbf{Feature} & \textbf{CORE-Bench} & \textbf{PaperBench} & \textbf{REPROBench} & {\sc \textbf{ReplicatorBench (Ours)}} \\ \midrule
Domain & Computer science, Biology, and Social sciences & Machine Learning & Social and Behavorial Sciences & Social and Behavorial Sciences \\ \midrule
Paper Source & CodeOcean & ICML 2024 & Journals, preprints, Twitter/X, Retraction Watch & Peer-reviewed journals \\ \midrule
\# Papers & 90 & 20 & 112 & 39 \\  \midrule
Nature of Work & Reproducibility & Reproducibility & Reproducibility & Replicability \\ \midrule
Task design \&  evaluation & Outcome-oriented & Outcome-oriented &  Outcome-oriented & Both process- and outcome-oriented  \\ \midrule
Labels & Single label: reproducible & Single label: reproducible  & 4 reproducibility levels (4 levels from \textit{irreproducible} to \textit{fully reproducible})  & Binary (\textit{criteria met} and \textit{criteria unmet}) \\ \midrule
Input & Code, original data, paper & Paper & Paper, original data, code, documentations, findings list & Paper, focal claim \\  \midrule
\# Tasks & 270 & 8,316 & 112 & 3,128 \\ \midrule
Evaluation & Numerical range check & LLM scorer, hierarchical aggregation & Comparison against manual ground truth & 3-stage aggregated scores (Auto + LLM) \\ \midrule
Evaluated LLMs & \lstinline{GPT-4o}, \lstinline{GPT-4o-mini} & \lstinline{O3-mini-high}, \lstinline{GPT-4o}, \lstinline{Gemini-2.0}, \lstinline{DeepSeek-R1}, \lstinline{o1}, \lstinline{Claude-3.5} & \lstinline{GPT-4o} & \lstinline{GPT-4o}, \lstinline{GPT-5}, \lstinline{GPT-5-mini}, \lstinline{o3} \\ \midrule
Internet access & Yes & Yes & No & Yes \\ 
\bottomrule
\end{tabularx}
\caption{Comparison of reproducibility benchmarks.}
\label{tab:benchmarks}
\end{table*}

\subsection{Task Design}
\label{sec:task-design}

{\sc ReplicatorBench} defines replication tasks for a research paper by mimicking human experts' replication process. 
The goal of a replication study is to assess whether the focal claim can be verified when tested on a new, similar dataset obtained independently from the dataset used in the research paper. 
We decompose this workflow into three stages: 
(1) \textbf{Extraction}, in which the agent ``reads'' the manuscript and extracts relevant information, such as method description and evidence, based on the focal claim, and then retrieve new data from the Internet for replication;
(2) \textbf{Generation}, in which the agent prepares the replication by completing a preregistration of the replication plan, setting up an executable environment, executing the replication codes, and producing the intermediate results; the results of this stage include intermediate and final raw output that is ready to be further analyzed and interpreted in the third stage;
and (3) \textbf{Interpretation}, in which the agent locates the agent's replication output, extracts relevant results, derives directly comparable results (if needed), compares them against the preregistered criteria and draw the conclusions.

\subsubsection{Extraction}
Given the focal claim and the original paper in a PDF format, the agent is expected to perform two subtasks in this stage. 
First, the replication process starts with gathering key information that could be used in the replication process about the claim.
Particularly, the agent is tasked with producing a structured document with extractions of $24$ gradable information pieces surrounding \textit{five} dimensions: \textit{focal claim}, \textit{data}, \textit{method}, \textit{results}, and \textit{metadata}. 
We refer to this document as the \textit{post-registration} of the \textit{original study}.
Next, the agent is tasked with searching for the replication data sample. In our benchmark, we model this process as an information retrieval task by requiring an agent to identify the appropriate replication data from the Web.
The output of the extraction stage contains the \textit{post-registration} document and a list of URLs for collecting replication data.

\subsubsection{Generation}
This stage addresses the computational aspect of the replication process that focuses on data processing and modeling. When humans replicate research claims, they often encounter a situation in which they do not have access to the original code or the original code was not executable due to compatibility or other reasons. To mimic this situation, we design an \emph{easy} setting in which both replication data and the code are available, and a \emph{hard} setting in which only the replication data is available.
Given either setting, we further break this stage into three steps.

\paragraph{Step 1: Design}
Mimicking the preregistration by human replicators, the agent is required to \textit{preregister} the research plan before carrying out the replication. 
This step involves documentation of hypotheses, data, and method planning, as well as preparation of computational environment. 
Under the easy setting, this step sets up the virtual environment by installing the chosen programming language and necessary software dependencies, as well as resolving visible issues such as non-supported programming scripts or filename inconsistencies.
Under the hard setting, the agent is expected to generate code following the original study's methodology description. 
The output of this step is a structured \textit{preregistration} document with 30 gradable information pieces that reflects the plans for the replication across \textit{seven} dimensions: \textit{hypothesis}, \textit{study type}, \textit{data plan}, \textit{proposed methodology}, \textit{codebase}, \textit{environment specifications}, and \textit{analysis steps}.


\paragraph{Step 2: Execution}
In this step, the agent performs the replication plan made in the last step. 
This includes processing the replication data, building models, deriving intermediate results, and generating raw results that could be later used for interpretation.
This step also includes iteratively debugging software issues at compilation and run time toward a successful execution of the code (either the original or re-implemented).
The output of this step is a structured document that reports how well the replication study is aligned with its preregistration, focusing on execution summary, intermediate results, and non-textual outputs such as tables and figures.

\renewcommand{\arraystretch}{1.0} %
\begin{table}[t]
\centering
\begin{tabular}{lc||lc}
\hline
Economics & 10 & Political Science & 4 \\
Education & 2 & Psychology \& Cognitive Science & 8 \\
Public Administration & 1 & Health & 9 \\ Sociology & 5 \\
\hline
\multicolumn{4}{c}{(a) Topics of 39 Papers.} \\
\end{tabular}
\begin{tabular}{cc|cc|c|cc}
\hline
\multicolumn{2}{c|}{Extract} & \multicolumn{2}{c|}{Generate} & Interpret & \multicolumn{2}{c}{Outcome} \\
Info & Web & Design & Exec. &  & {\small Criteria} Met & Unmet \\
936 & 86 & 1170 & 390 & 507 & 20 & 19 \\
\hline
\multicolumn{7}{c}{(b) Distributions of 3,128 Tasks.} \\
\end{tabular}
\caption{Statistics of {\sc \textbf{ReplicatorBench}}. All tasks were created from the entire sample of 39 papers, with the exception of the Extract-Web substage, which applies only to the 19 observational studies whose replication data is available on the web.}
\label{tab:data-stats}
\vspace{-0.3in}
\end{table}



\subsubsection{Interpretation}
The goal of the Interpretation stage is to assess the outcome of a replication attempt, taking as input the execution output and logs, along with all inputs and outputs associated with the previous stages. 
First, the agent inspects the execution logs to checks what was executed versus what was planned in the design step.
Next, it compares the final results of the execution step to the original study results recorded in the post-registration; verifying key reported statistics and evaluating the hypothesis tied to the focal claim.
Third, the agent is required to determine whether the inference criteria for the focal claim are satisfied (i.e., a statistically significant effect ($\alpha = 0.05$, two-tailed) in the same pattern as the original study) and draw the final conclusion.
The output of this stage is a structured document with 13 gradable result interpretations across \textit{seven} dimensions: \textit{summary}, \textit{fidelity assessment}, \textit{results comparisons}, \textit{replication report}, and \textit{execution failure handling}.
We provide an overview of the task templates and supplementary artifacts in Appendix \ref{sec:task-templates}.


\subsection{Rules}

Through the replication workflow, an agent has read and write access to all files in the designated workspace on a dedicated computer. 
The agent has the permission to install, uninstall, and execute software packages. 
The agent works on the replication task in a closed environment, except for the Extraction stage, in which they are provided with Internet access. 
{\sc ReplicatorBench} technically does not require any human intervention. 
However, as a precaution, we implemented an optional human-check function for all executable commands and file-writing actions. 
When enabled, actions are executed only after approval by a human supervisor.
We observed no significant unsafe actions when testing the agent.

\section{Dataset}
\label{sec:dataset}

{\sc ReplicatorBench} evaluates LLM agents on replication tasks over a set of focal claims, each identified by domain experts from the papers compiled by the SCORE project \cite{alipourfard2021systematizing}. 
These papers are sampled from peer-reviewed journals in six subject categories in social and behavioral sciences \cite{alipourfard2021systematizing}. 
From the broader set of SCORE replications, we keep only papers where (1) the focal claim can be tested with a single quantitative check and a clear pass/fail criteria, and (2) the core materials needed for replication are available (preregistration/replication plan, data access, and code documentation). 
We do not restrict the programming languages used in the replication code.
Since many SCORE cases are missing at least one of these pieces, this filtering yields 39 studies. Of these, 20 are observational in which the replication sample can be built from online or archival sources, and 19 are experimental studies in which the replication team must collect data through controlled experiments.

For each paper, a team of human researchers conducts a replication study to assess the replicability of the focal claim using a new dataset. 
The human replicators first pre-register an analysis that specifies data construction, models, and decision criteria used to judge whether the claim replicates. For each claim, the human replication team delivers a replication report documenting the preregistered plan, how the new data were obtained and processed, instructions to execute scripts, intermediate results, and the final conclusion based on whether the \emph{preregistered criteria for a successful replication are met} (e.g., an effect that is statistically significant and in the same direction as the original study).
These reports provide the ground truth for evaluating the agents. 
By breaking down the replication workflow of each claim into three stages, the benchmark contains 3,128 task instances (Table \ref{tab:data-stats}).

\subsection{Grading}

We pair each stage (see $\S$\ref{sec:task-design}) with manual documentation from a human replication effort to evaluate the performance of the agent.

\paragraph{Reference for Extraction} To assess the quality of the agent's extracted information,
we recruit three human annotators, all with research experience, to annotate references for the extraction tasks based on the original paper and focal claim.
We ensured all three annotators passed a test to annotate relevant information from a representative paper before performing a complete annotation of the 39 papers.
We adopt a leave-one-out evaluation setting where we use extracted information from two annotators as the ground-truth reference and the extracted information by the third annotator for computing a baseline performance for the extraction stage.

The ground truth of the data retrieval task is the URLs linking to the new data documented by human replicators in the preregistration document. 
Agents are not provided any candidate URLs. 
Given the original paper and the claim to replicate, an agent must perform an open web search to locate the appropriate data source.

\paragraph{Reference for Generation-Design}
To assess the quality of the agent's preregistered plan for the replication study, we also use the original preregistration document by the human researchers. 
Specifically, we rely on the entire preregistration document, which include reported data cleaning, preprocessing procedures, and proposed methodologies in a single PDF or DOC file, to measure how well-developed the agent's own plans are. 

\paragraph{Generation-Execution Checkpoints} As the generation stage is mainly concerned with the computational capabilities of the agent, we also identify 10 reference-free checkpoints for each claim on which the agent is evaluated in this stage. 
The \textit{computational design} checkpoints include setting up the environment, identifying package dependencies, and investigating the file system; the \textit{execution} checkpoints assess code execution and debugging capabilities, as well as appropriate documentation of execution outputs for the interpretation stage. 
The checkpoint schema used for execution evaliation is included in the supplementary artifact~\cite{artifact}.


\paragraph{Reference for Interpretation}
The final report produced by the human replication team, in a single PDF or DOC file, is used as a reference for the Interpretation stage.

\section{Evaluation Metrics}
\label{sec:evaluation}

Despite requiring agents to format their responses in a structured template, LLM-generated texts remain open-ended and are sensitive to reference-based metrics like BERTScore or ROUGE \cite{zhang2019bertscore,lin-2004-rouge}. 
We compare the effectiveness of LLM-as-a-judge (\textit{LLMEval}) \cite{zheng2023judging} with other text-generation metrics in Appendix \ref{sec:app-evaluation} and found that LLMEval can provide reliable judgment of agent's replication attempts against the reference in our benchmark.
Thus, we adopt LLMEval for automatic evaluation of extracted and generated results in all checkpoints as follows:

\setdefaultleftmargin{1em}{2em}{}{}{}{}
\begin{compactitem}
\item{\textit{Extraction}:
    For each piece of information, the LLM judge assigns a score in a 0-3 range to measure how well extracted information matches the reference, with 0 indicating a poor match with the reference and 3 indicating a perfect match. 
    The final score of this stage is calculated as the macro average of all 24 extracted fields, distributed across the five dimensions discussed in $\S$\ref{sec:task-design}.}
    \item{\textit{Generation - Design}: The LLM judge directly uses the human-submitted preregistration document, in a PDF or DOC format, to evaluate the quality of the agent-generated preregistered plans. 
    For each field in the generated plan, the judge assigns a score of 0 or 3, with 0 indicating a poor match with the reference and 3 indicating a perfect match. 
    The final score of this stage is calculated as the macro average of 30 generated fields, distributed across seven dimensions discussed in $\S$\ref{sec:task-design}.}
    \item{\textit{Generation - Execution}: The LLM judge inspects the logs and execution outputs in the design and execution steps to determine whether the agent has addressed all 10 generation checkpoints. 
    The LLMEval assigns a score of 0 or 1 for each generation checkpoint, with 0 indicating an unsatisfactory attempt while 1 indicates a satisfactory attempt. 
    The final score of this stage is calculated as the macro average of 10 checkpoints across two dimensions \textit{computational design} and \textit{execution}.}
    \item{\textit{Interpretation}: The LLM judge compares the agent's interpretation directly against the human replication report.
    For each checkpoint instance in this stage (e.g., statistical results and overall conclusion), the LLMEval assigns a score in a range of 0 to 3 for semantic match, with 0 indicating a poor match with the reference and 3 indicating a perfect match.
    The final score of this stage is calculated as the macro average of 13 generated fields, distributed across seven dimensions discussed in $\S$\ref{sec:task-design}.}
\end{compactitem}

In addition to reporting rubric-based scores with LLMEval, we also report binary classification metrics for the final checkpoint in which we compare the agent's conclusion using the human researchers' conclusion (i.e., whether they both agree on \textit{Criteria Met} or \textit{Criteria Unmet}). 
We observe that it is possible that the agent makes an inconclusive decision when it is unable to execute the code successfully or when it fails to explore the execution outputs during the interpretation stage. 
We treat such decisions as incorrect decisions. 
We report precision, recall, and F1 scores for this checkpoint. 
Our benchmark contains 20 human replication studies labeled as \textit{criteria met} and 19 labeled as \textit{criteria unmet}.
To treat these two classes equally, we use macro aggregation to reflect agents' ability to assess both replicable and non-replicable claims. 

For the data search task in the Extraction stage, we evaluate the retrieved URLs with alias matching. 
Because resources may have multiple entry points and landing pages that require human expert

\noindent authentication, we treat the model's output as correct if it matches any of the aliases. 
We report macro precision, recall, F1, hit@\textit{any}, and hit@\textit{all}, with both micro and macro aggregation across cases. 
Hit@\textit{any} measures the frequency with which the search model found at least one required resource from the gold set, and hit@\textit{all} measures if all required resources were found.

\begin{figure}[t]
    \centering
    \includegraphics[width=\linewidth]{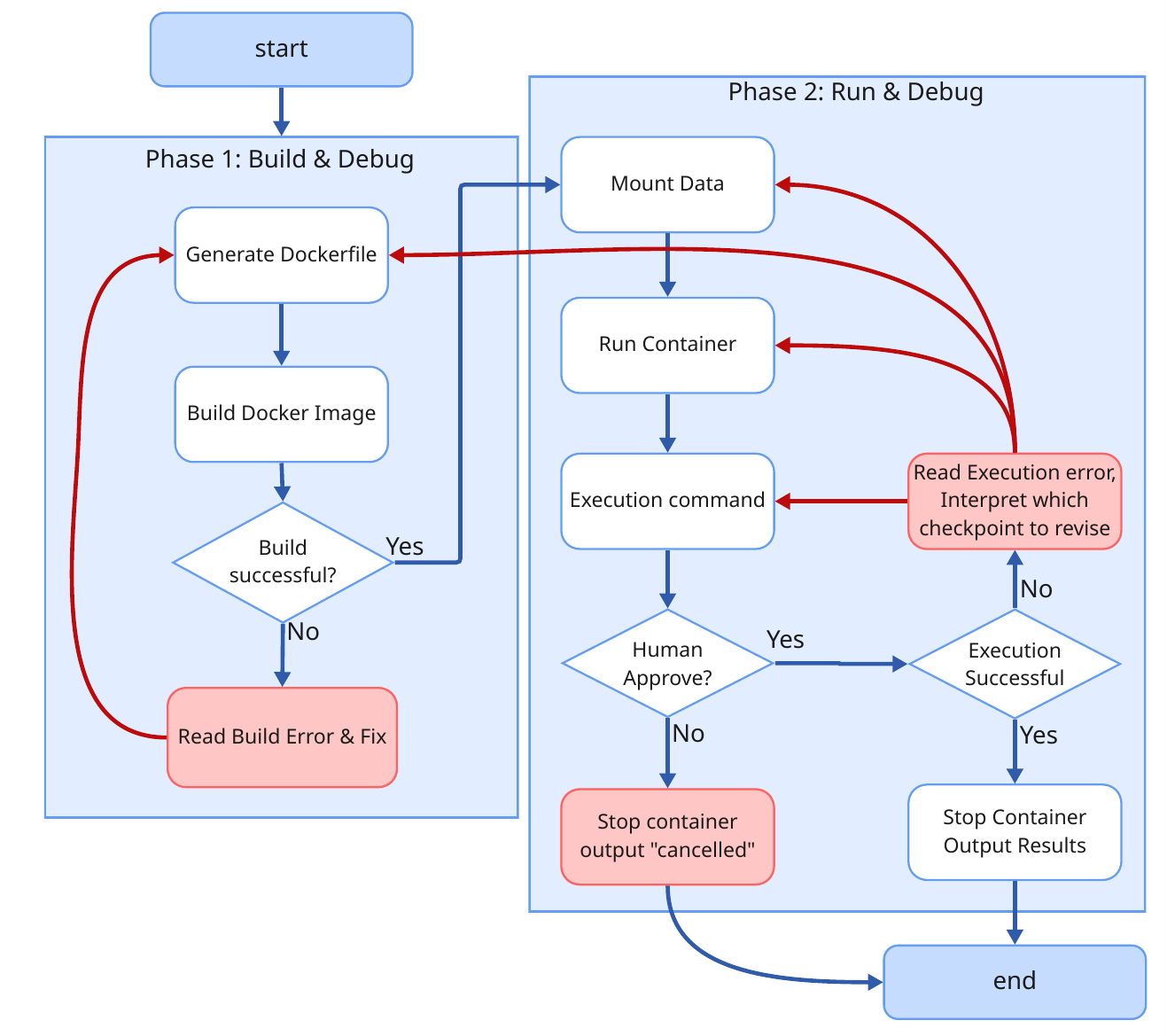}
    \caption{The Generation-Execution stage has two main phases: build and debug, and run and debug. We developed iterative debugging to fix issues and improve performance.}
    \label{fig:iterative_debuggging}
    \Description[Figure illustrating iterative debugging]{The Generation-Execution stage has two main phases: build and debug, and run and debug. We developed iterative debugging to fix issues and improve performance.}
\end{figure}

\section{Experimental Setup}
\label{sec:experiments}
As a baseline study, we developed an agentic framework called {\sc ReplicatorAgent} to demonstrate the performance of frontier LLMs against our benchmark. 

\subsection{{\sc \textbf{ReplicatorAgent}}}
{\sc ReplicatorAgent} follows a ReAct-style loop \cite{yao2022react}, in which it alternates between reasoning and actions to execute the replication workflow and produce the required artifacts in each stage, such as the preregistration and post-registration described in $\S$\ref{sec:dataset}, and execution logs. 
One key design consideration is to support iterative debugging rather than treating the first execution failure as terminal (Figure~\ref{fig:iterative_debuggging}).
We include an example execution log showing this iterative debugging in practice in Figure~\ref{fig:scipy}.

The {\sc ReplicatorAgent} is provided with a simple palette, which was motivated by the commonly experienced replication failures we experienced. 
This palette is effective in reducing single-point failures such as hard-coded paths, missing dependencies, and partial outputs. 
The palette includes file and directory inspection, targeted file readers, dataset inspection utilities such as column names and basic summaries, file editing that produces git-style diffs for localized patches, and constrained rewrites that require explicit overwrite flags to reduce destructive edits.

\subsection{Execution Environment} 
We run the agentic framework in a sandboxed container environment orchestrated by a runner, a lightweight orchestration script that builds an image, mounts the code and data into a standard directory layout, executes the analysis entrypoint, and exports the logs and other artifacts produced.
Sandboxing the environment serves as a safety measure and ensures that our results are reproducible by recording the exact environment and executed commands.

As agents write and execute code in the environment, raw execution logs may exceed context limits and alter the statistics needed for the final criteria decision. 
In these cases, we truncate the logs by keeping the first 2000 lines and then prompt the agent to rewrite the analysis entrypoint to produce a compact summary and rerun the analytics so that key information prompted early on, such as the preregistered coefficients and $p$-values are included within the context limit. 
This practice produces compact outputs that are used by the \emph{Interpretation} stage without fabricating missing evidence. 

\subsection{Design Choices}
\paragraph{Native vs. Python Modes}
Inspired by existing findings regarding LLMs' preference for Python during code generation and execution \cite{twist2025llms} and thereby superior performance in Python coding compared to other languages, e.g., R \cite{siegel2024core}, we evaluate two execution configurations:
\textit{Native} mode, in which the agent runs the replication package in its original programming language when possible and is explicitly instructed not to translate code;
and \textit{Python} mode, in which the agent is required to translate a non-Python script into Python prior to setting up the execution environment. 


\paragraph{Data-only Setting}
Inspired by the real-world challenge, we test the agent under two possible settings, depending on whether the code is accessible in the replication package. In both settings, the agent has access to the new data (the agent is still required to accomplish the data retrieval task). However, in the data-only setting, the agent does not have access to the native codes. To accomplish the replication tasks, the agent must generate its own implementation of the methods based on the algorithms conveyed in the paper. This ablation study tests the agent's capability to implement methods solely by reading documentation and extracted specifications. 

\paragraph{Underlying LLMs}
We evaluate our framework by instantiating its policy with multiple frontier LLMs.
Specifically, we experimented with GPT-4o, GPT-5, GPT-5-mini, and o3, which differ in reasoning strength and tool use behavior.
All models are evaluated under the same agent architecture, tools, and execution settings to draw a fair comparison.

\section{Results}

\begin{table*}[t]
\centering
\resizebox{\textwidth}{!}{%
\begin{tabular}{l|c|ccccc|c|c|c|cccc}
\toprule
\multirow{3}{*}{\textbf{Model}}
& \multicolumn{6}{c|}{\textbf{Extract}} 
& \multicolumn{2}{c|}{\textbf{Generate}} 
& \multicolumn{1}{c|}{\multirow{2}{*}{\textbf{Interpret}}}
& \multicolumn{4}{c}{\multirow{2}{*}{\textbf{Outcome}}} \\
\cmidrule(lr){2-7} \cmidrule(lr){8-9} 
& \makecell{\textit{Info}} & \multicolumn{5}{c|}{\makecell{\textit{Web}}} 
& \makecell{\textit{Design}} & \multicolumn{1}{c|}{\makecell{\textit{Execute}}} 
&  \multicolumn{1}{c|}{\makecell{\textit{}}} & & \\
& LLMEval & Prec. & Rec. & F1 & hit@\textit{any} & hit@\textit{all} & LLMEval & LLMEval & LLMEval & Acc. & Prec. & Rec. & F1 \\
\midrule
\lstinline{o3}         & 60.96 & 13.59 & 28.60 & 16.68 & 57.89 & 15.79 & \textbf{83.82} & \textbf{95.18} & \textbf{94.54} & 66.67 & 67.06 & 67.07 & 66.67 \\
\lstinline{GPT-4o}    & 57.56 & \textbf{21.75} & 22.53 & \textbf{19.49} & 57.89 & 10.53 & 77.45 & 80.48 & 82.25 & 61.54 & 61.45 & 61.51 & 61.44 \\
\lstinline{GPT-5-mini} & 62.93 & \ \ 7.09  & 28.81 & 10.56 & 57.89 & \textbf{15.79} & 69.20 & 69.07 & 87.02 & 48.72 & 50.50 & 50.50 & 47.86 \\
\lstinline{GPT-5}     & \textbf{63.65} & \ \ 7.41  & \textbf{30.62} & 10.95 & \textbf{63.16} & \textbf{15.79} & 79.38 & 95.14   & 91.26 & \textbf{76.92} & \textbf{78.53} & \textbf{77.78} & \textbf{76.86} \\
\bottomrule
\end{tabular}%
}
\caption{Performance of our LLM agents across different stages against {\sc \textbf{ReplicatorBench}} in Python mode. GPT-4o is used as the backbone model for LLMEval. All evaluation scores are normalized to a 0-100 scale. Macro aggregations are used.} 
\label{tab:merged_results}
\end{table*}

\begin{table*}[t]
\centering
\small
\resizebox{\textwidth}{!}{%
\begin{tabular}{l|l|rrr|rrr|rr}
\toprule
\textbf{Setting} & \textbf{Model} &
\textbf{Macro P} & \textbf{Macro R} & \textbf{Macro F1} &
\textbf{Micro P} & \textbf{Micro R} & \textbf{Micro F1} &
\textbf{Hit@\textit{any}} & \textbf{Hit@\textit{all}} \\
\midrule
\multirow{3}{*}{Search-tuned}
& o3-deep-research      & \textbf{25.35} & 22.09 & \textbf{23.26} & 18.57 & 15.85 & 17.11 & 52.63 & 10.53 \\
& gpt-5-search-api      & 7.19  & 11.58 & 7.82  & 10.34 & 3.66  & 5.41  & 15.79 & 10.53 \\
& gpt-4o-search-preview & 9.56  & 11.21 & 8.82  & 9.21  & 8.54  & 8.86  & 26.32 & 5.26 \\
\midrule
\multirow{4}{*}{Base + tool}
& gpt-4o                & 21.75 & 22.53 & 19.49 & \textbf{21.54} & 17.07 & \textbf{19.05} & 57.89 & 10.53 \\
& gpt-5                 & 7.41  & \textbf{30.62} & 10.95 & 6.88  & \textbf{23.17} & 10.61 & \textbf{63.16} & \textbf{15.79} \\
& gpt-5-mini            & 7.09  & 28.81 & 10.56 & 6.90  & 17.07 & 9.82  & 57.89 & \textbf{15.79} \\
& o3                    & 13.59 & 28.60 & 16.68 & 12.39 & 17.07 & 14.36 & 57.89 & \textbf{15.79} \\
\bottomrule
\end{tabular}%
}
\caption{Web search subtask performance across search-tuned models and base models equipped with web search tools. We report macro and micro precision (P), recall (R), and F1, along with hit@\textit{any} and hit@\textit{all}. All values are scaled to a 0--100 range.}
\label{tab:websearch}
\end{table*}

\subsection{Main Results}
Table~\ref{tab:merged_results} presents the performance of {\sc ReplicatorAgent}, initiated with different LLMs, in the three stages of {\sc ReplicatorBench} in \textit{Python mode,} where all non-Python scripts are translated before execution to keep the environment consistent across models.
We report the \textit{Native} vs. \textit{Python} comparison separately.
This control for differences in language-specific runtimes helps isolate the LLM's bias towards certain programming languages.

\subsubsection{Extracting Claim Information}
Under leave-one-out evaluation, our human annotators achieve an LLMEVal score of $71.33$ on information extraction. 
GPT-5 achieves the highest score ($63.65$) and GPT-4o achieves the lowest score ($57.56$) for this task.
All evaluated LLMs have yet to measure up to human performance in filling out the post-registration.

\subsubsection{Web search for replication data}
Because of the noticeable performance difference between the data search and other tasks, we compared base models and several fine-tuned models on the data retrieval search task. 
The results in Table~\ref{tab:websearch} indicate that among fine-tuned search models, \model{o3-deep-research} achieves the strongest overall performance 
and the highest coverage 
while \model{gpt-4o-search-preview} has higher hit@\textit{any} than \model{gpt-5-search-api} 
, and \model{gpt-5-search-api} is higher on hit@\textit{all}.
Interestingly, tool-augmented base models exhibit substantially higher coverage overall 
, but this comes with differing precision–recall tradeoffs: \model{GPT-4o} has the highest precision and F1
, whereas \lstinline{GPT-5} achieves higher recall and the best hit rates 
despite lower precision. 
Base models benefit from stronger long-context reasoning and iterative query refinement, which are critical for mapping research claims to dataset sources, whereas search-tuned models may optimize generic web relevance and thus return plausible but non-canonical pages that are penalized under domain-based grading.

\subsubsection{Computational experimentation}

The evaluation provided by LLMEval indicates that o3 and \lstinline{GPT-5} have the strongest computational performance for the execution stage. 
However, o3 exhibits a better performance during the preceding design stage that plans for the replication study compared to \lstinline{GPT-5}.
Upon manual inspection, we found that \lstinline{GPT-5} tends to shift effort from design to the execution stage.
For example, the model only generates planning steps for the translation of the given code into Python during the \textit{design} step and then actually performs the data inspection and code translation in the \textit{execution} step.
This difference in where the translation work is surfaced (Design vs.\ Execute) aligns with the stage-wise LLMEval rubric: designs that are more explicit and immediately actionable score higher, which explains why o3 achieves a better Design score than \lstinline{GPT-5}. 
This is not a substantive failure of \lstinline{GPT-5}, but rather a different task decomposition strategy.

\subsubsection{Results interpretation and replication outcomes}

Given an effective performance during the execution stage, our \lstinline{GPT-5} and o3 agents arrive at conclusions in the Interpretation stage with LLMEval scores of 91.26 and 94.54, respectively, compared to 82.25 and 87.02 by \lstinline{GPT-4o}, and \lstinline{GPT-5-mini}. 
Regarding binary classification performance on replication outcome, \lstinline{GPT-5} achieves the best performance across all LLMs, with a $10\%$ net gain compared to the runner-up \lstinline{o3}. 
\lstinline{GPT-5-mini} has yet to demonstrate competence for this task, only achieving a Macro F1 of $47.86$.

\subsection{Additional Analyses}


\subsubsection{Native vs Python}
\begin{figure}[t!]
    \centering
    \includegraphics[width=0.45\textwidth]{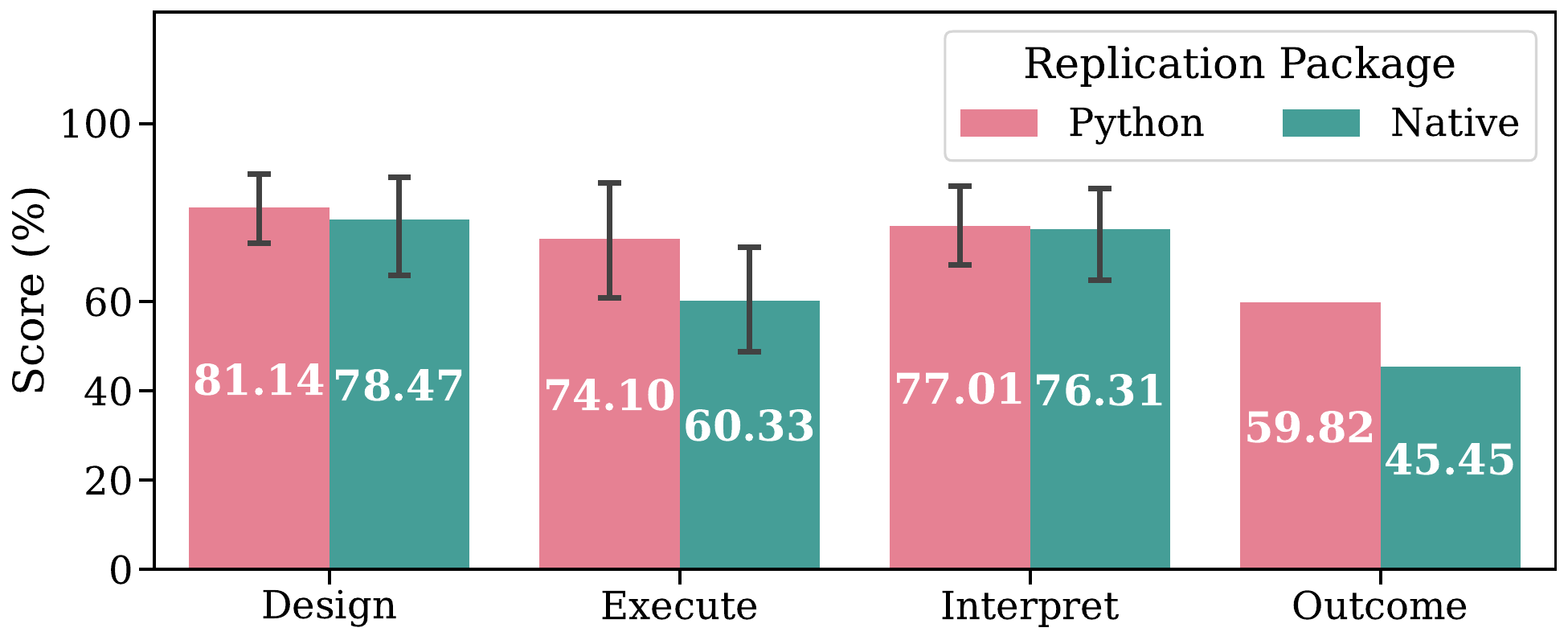}
    \caption{Performance of GPT-4o {\sc \textbf{ReplicatorAgent}} in a Python-only setting compared to Native setting. Bars show LLMEval stage scores with error bars calculated as a 95\% confidence interval of the mean. Outcome reports macro F1.}
    \label{fig:native-vs-python}
    \Description[Barplot of comparing Native vs Python configurations]{Performance of GPT-4o {\sc \textbf{ReplicatorAgent}} in a Python-only setting compared to Native setting. LLMEval scores are reported for design, execution, and interpretation stage with error bars calculated as a 95\% confidence interval of the mean. Macro F1 Scores are reported for the final replication outcome (criteria met or unmet)}
\end{figure}

In this analysis (see Figure~\ref{fig:native-vs-python}), we highlight that an agent's capability for computational experimentation may not necessarily correlate with their capability to sensibly arrive at the correct replication outcome. 
Results indicate that Python mode often improves the agent's capability to reach an executable run when the native codes fail due to dependency installation or environment compatibility issues.
Although this practice increases the performance of the execution stage, it does not necessarily guarantee the reliability of the final assessment regarding the focal claim's replicability. 
We attribute this phenomenon to potential hallucinations and information loss as the agent translates original code into Python.
Incorrect code translations may lead to different numerical outputs compared to the execution of the original code. These errors are then propagated to the interpretation stage.
We provide an example of such errors in Figure~\ref{fig:translation-error} in Appendix~\ref{sec:qualanalysis}.
Figure~\ref{fig:native-vs-python} highlights the results for GPT-4o in this analysis, and we observe similar trends in GPT-5 and o3 (see Table \ref{tab:native-vs-python} in Appendix ~\ref{app:additional-results}).

\subsubsection{Data-only Setting}
\begin{figure}[t]
    \centering
    \includegraphics[width=0.45\textwidth]{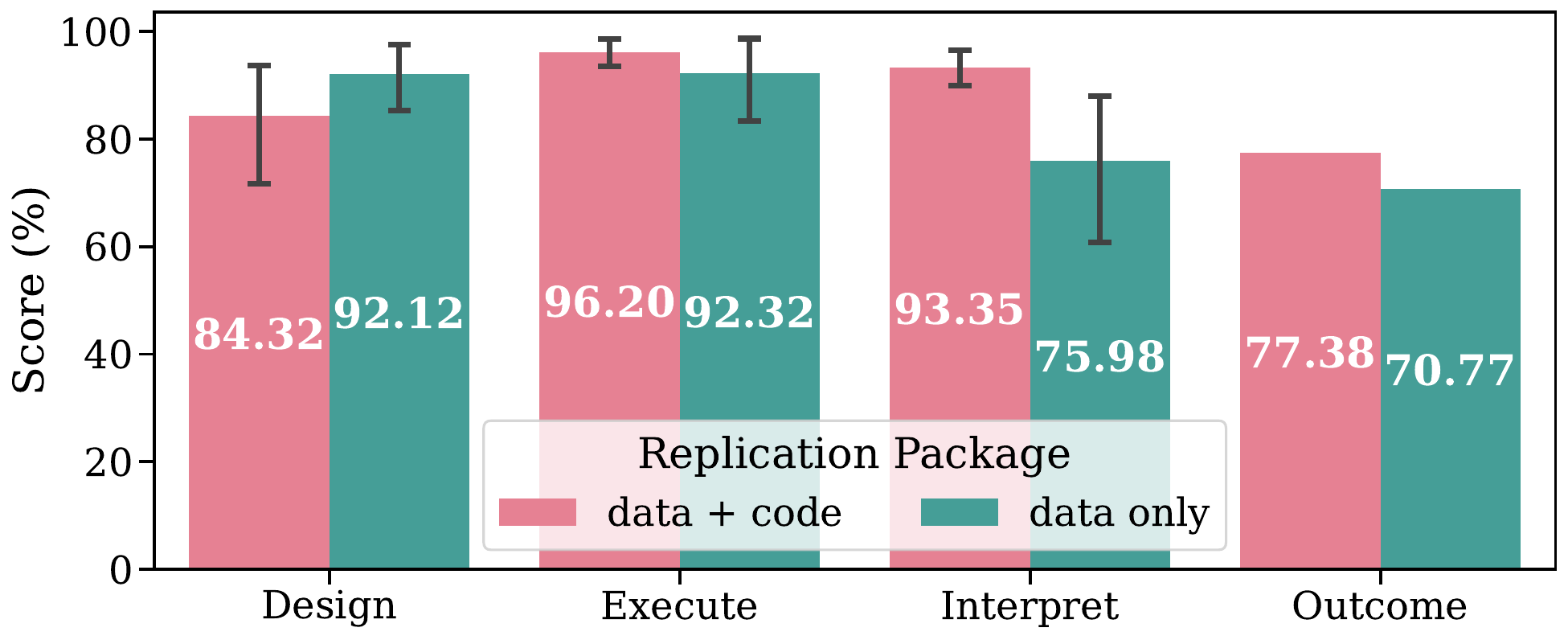}
    \caption{GPT-5 {\sc \textbf{ReplicatorAgent}} with data+code versus data-only access. Bars show LLMEval stage scores with error bars calculated as a 95\% confidence interval of the mean. Outcome reports macro F1.}
    \label{fig:data-only-analysis}
    \Description[Performance barplot for GPT-5 comparing with and without code for each stage]{Performance of GPT-5 {\sc \textbf{ReplicatorAgent}} with and without access to human-written replication code on the 19 observational studies. LLMEval scores are reported for design, execution, and interpretation stage with error bars calculated as a 95\% confidence interval of the mean. Macro F1 Scores are reported for final replication outcome (criteria met or unmet).}
\end{figure}

Figure~\ref{fig:data-only-analysis} compares the performance of GPT-5 {\sc ReplicatorAgent} with and without access to human-written replication code. 
We observe an improvement in preregistering the research plan (higher LLMEval score for the design stage) when the agent has to write and execute code on its own. 
Under this setting, the agent typically spends more turns exploring the data to help generate the replication code, in turn providing a more detailed data plan for the preregistration. 
However, as the agent has to navigate both computational performance (generating and executing codes) and fidelity to original methodology in the generated code, we observe a drop in the agent's alignment with human researchers in their final assessment of the claim's replicability (lower scores for interpretation stage and replication outcome). 

\subsection{Error Categories}

To support error analysis, we define the following error categories and use them to label errors when the replication is \emph{Partial Success} or \emph{Failure}. A single run may be assigned multiple error labels.
\begin{compactitem}
\item{\textit{Setup errors}. The run cannot be started due to environment or dependency problems, e.g., installation errors, or missing files.}
\item{\textit{Input data errors}. The dataset cannot be loaded or does not match what the analysis expects, e.g., corrupted inputs, missing required variables, or incompatible formatting.}
\item{\textit{Implementation errors}. The run does not execute the intended procedure, e.g., omitting a required variable, incompatible variable types, or implementing a different procedure.}
\item{\textit{Result extraction errors}. The run does not produce the required outputs, e.g., target quantities cannot be located, or expected output files are missing.}
\item{\textit{Timeout errors}. The run does not finish within the allowed time, or the agent reaches its iteration limit.}
\end{compactitem}

We analyzed the distribution of error types across all models to identify potential bottlenecks in replication (see Appendix~\ref{sec:error-analysis} for details). 
Our analysis reveals that Input data errors, specifically in loading datasets and handling variable types, were dominant for GPT-4o. 
In contrast, newer models like GPT-5 and o3 demonstrate higher stability with fewer environment and setup crashes, reducing the total volume by nearly 80\% and effectively mitigating environment and data handling issues. 


\section{Discussion and Future Work}
Our results highlight the strengths and weaknesses of LLM agents in replicating research claims in SBS domains.
Across the models and execution settings, agents demonstrate strong capabilities in structured code execution and debugging. 
However, they exhibit substantially weaker performance in earlier stages that require locating replication data on the Internet.
Importantly, successful execution does not always translate into correct interpretation.
Even when agents are able to reach beyond the generation stage and produce numerical results, interpretation errors, deviations from the pre-registered plan, and subtle implementation differences can lead to incorrect decisions.
This gap suggests that benchmarks focusing solely on execution success may overestimate an agent's real-world utility, highlighting the importance of evaluating multiple stages of the replication process rather than a single outcome. 

{\sc ReplicatorBench} highlights promising directions for the development and evaluation of AI research assistants.
Because current models struggle to locate appropriate data resources, we call for future work on agent development, specifically focusing on the planning stages, developing more specialized tools and effective search strategies to construct new replication samples.
Additionally, as {\sc ReplicatorBench} focuses on observational studies in the SBS domains where data is web-retrievable, future work should develop benchmarks for experimental replication.
In these scenarios, agents must navigate controlled settings to collect or generate primary data.
Together, these directions move the literature toward autonomous, end-to-end AI assistants for research replication.

\section{Limitations and Ethical Considerations} 
\paragraph{Limitations}
We acknowledge several limitations of {\sc ReplicatorBench}.
First, the benchmark is constructed from a sample of 39 replication studies.
This scale is constrained by the scarcity of high-quality, expert-documented replication efforts that span multiple research stages.
For the extraction stage, we collected three expert-annotated extractions about the focal claim of each study and documented necessary web resources.
For the generation and interpretation stages, a group of researchers independently conducted a replication for each study. 
These efforts followed the standard scientific process: preregistering a research plan, conducting the study, and submitting final reports alongside a replication package of code and data.
{\sc ReplicatorBench}, thus, represents a deliberate design choice to prioritize data quality and process fidelity.
We also recognize the shortcomings of using LLM-as-a-judge (LLMEval) for grading open-ended text, treating the rubric score as approximations rather than an absolute measure of replication competence. 
To mitigate the uncertainty of LLMEval, we used two complementary evaluation strategies. 
First, in extraction, we use LLMEval to assess our annotators' extractions in a leave-one-out setting to establish an upper bound. 
Second, we provide binary classification metrics for replication outcomes, which exhibit assessment patterns similar to LLMEval in the interpretation stage.

\paragraph{Ethical Considerations}
We address several ethical considerations regarding data privacy, consent, bias, and potential misuse.
Regarding data privacy and consent, all studies and replication packages in {\sc ReplicatorBench} were sourced from the SCORE program with appropriate licenses (see $\S$\ref{sec:dataset}).
We ensured data diversity in our benchmark by selecting 19 papers from six disciplines in the social sciences (see Table~\ref{tab:data-stats}).
Regarding potential misuse, we recognize the risks of allowing LLM agents to autonomously write and execute code. 
We addressed this concern by setting up a sandboxed environment for each run and each study to isolate agent actions from host systems (see $\S \ref{sec:experiments}$). 
We also adopted a human-check procedure for all executable commands and file-writing actions and observed no significant unsafe actions from the agents.

\section{Conclusion}

We introduced {\sc ReplicatorBench}, a benchmark for evaluating LLM agents' capabilities to replicate research claims in SBS domains. 
Unlike prior benchmarks that focus mainly on execution under ideal conditions, {\sc ReplicatorBench} breaks down the workflow into multiple stages, mimicking the human replication workflow. 
We grounded each stage in expert-led human replication efforts with a preregistered plan and final reports, which allows us to evaluate an agent's behavior at each stage rather than relying on only the final outcome. 
We developed a baseline agentic framework, {\sc ReplicatorAgent}, and evaluated it against {\sc ReplicatorBench} under multiple underlying LLMs and configurations.
Our results show that although state-of-the-art LLMs are often capable of performing complex computational workflows and iteratively resolving execution failures, this performance does not consistently yield correct replication judgments. 
Furthermore, agents struggle in earlier stages, particularly in locating replication data on the Web, implying the importance of moving towards benchmarks that capture the entire replication workflow rather than just the outcomes. 



\begin{acks}
This project is funded by Coefficient Giving as part of its 'Benchmarking LLM Agents on Consequential Real-World Tasks' program. 
We thank Anna Szabelska, Adam	Gill, and Ahana Biswas for their annotation of the ground-truth post-registrations for the extraction stage.
\end{acks}

\bibliographystyle{ACM-Reference-Format}
\bibliography{_main}

\appendix

\section{Error Analysis}
\label{sec:error-analysis}

We categorize the execution failures into \textbf{Setup} (environment/dependency failures), \textbf{Input} (loading or formatting issues), \textbf{Implementation} (logic or syntax errors), and \textbf{Timeout} errors. 

\subsection{Quantitative Results}
To classify these errors, we developed an automated parsing script that processes the raw execution logs for each run for each model. 
The script utilizes regular expressions to detect Python tracebacks, R execution halts, and standard error (\textit{stderr}) streams within the agent's JSON observation fields
Errors are then mapped to categories based on keyword signatures.
While this heuristic approach allows for scalable analysis across the logs, it is an approximation.
Ambiguous errors, like a file missing because the previous step failed vs. it actually not existing, are categorized based on the immediate error signal.

As shown in Table~\ref{tab:error_dist}, \textbf{Input data errors} were the most prevalent category across all models, accounting for the vast majority of failures in GPT-4o runs ($N=124$).
Qualitative inspection suggests that these were primarily driven by the model's inability to robustly handle diverse file formats or strict datatype requirements in the provided datasets.
\textbf{Implementation errors} were the secondary cause for GPT-4o ($N=33$), but presented the primary bottleneck for the smaller GPT-5-mini ($N=74$), which struggles more with correct syntax and library usage than data handling.
Notably, GPT-5 and o3 achieved a massive reduction in total failures, with GPT-5 encountering only 6 Input data errors compared to GPT-4o's 211, highlighting a substantial improvement in the model's ability to navigate complex data environments without crashing.

\begin{table}[ht!]
\centering
\begin{tabular}{lcccc}
\toprule
\textbf{Error Category} & \textbf{GPT-4o} & \textbf{GPT-5} & \textbf{GPT-5-mini} & \textbf{o3} \\
\midrule
Setup Errors          & 35  & 21 & 4  & 9 \\
Input Data Errors     & 124 & 6  & 0  & 1 \\
Implementation Errors & 33  & 11 & 74 & 25 \\
Timeout Errors        & 1   & 0  & 0  & 0 \\
\midrule
\textbf{Total}        & 193 & 38 & 78 & 35 \\
\bottomrule
\end{tabular}
\caption{Distribution of runtime errors across models for 19 observational studies in {\sc \textbf{ReplicatorBench}}. GPT-4o failures are dominated by \textbf{Input Data Errors} (data loading, type mismatches), whereas GPT-5-mini primarily exhibits \textbf{Implementation Errors} (coding logic and syntax).}
\label{tab:error_dist}
\end{table}

\subsection{Qualitative Analysis} \label{sec:qualanalysis}
A key design feature of our {\sc ReplicationAgent} is \textit{iterative debugging}, which is the ability to recover from execution failures. 
We present two contrasting cases to illustrate the capabilities and limitations of this approach. 

\paragraph{Case 1: Successful Environment Recovery}

Figure~\ref{fig:scipy} demonstrates how GPT-4o was able to resolve a Docker failure caused by an incompatible library dependency.
Rather than terminating, the agent utilized the error trace to diagnose the specific version conflict, proposed a targeted edit to the configuration file (\textit{replication\_info.json}), and successfully rebuilt the Docker image to achieve a successful run.
This confirms the agent's ability to handle explicit errors.

\begin{figure}[htbp!]
\centering
\begin{minipage}{0.5\textwidth}
\begin{lstlisting}[style=logstyle]
---Turn 2 ---
Thought: The Dockerfile has been successfully generated.  The next step is to build the Docker image using this Dockerfile. 
Action: orchestrator_build_image: "./data/original/4/python"
PAUSE
Running Action: orchestrator_build_image [Checkpoint: 2. Build Image]

--- Turn 3 ---
Observation: Docker build failed. See 'build_log' for details (pip dependency mismatch)
Build_log (excerpt):
<<R>>ERROR: Ignored the following versions that require a different python version: ...<<END>>
<<R>>ERROR: Could not find a version that satisfies the requirement scipy==1.7.1<<END>>
... [built output truncated] ...
Thought: The Docker build failed because the specified version of `scipy` (1.7.1) is not compatible with the Python version in the Docker image. <<G>>I need to update the `replication_info.json` to specify a compatible version of `scipy` for Python 3.10. I will choose a version of `scipy` that is compatible with Python 3.10, such as `1.7.3`, which is available according to the error log.<<END>>
Action: read_file: "./data/original/4/python/replication_info.json"

--- Turn 4 --- 
Thought: I need to update the `replication_info.json` to specify a compatible version of `scipy` for Python 3.10. <<G>>Based on the error log, `scipy==1.7.3` is a compatible version. I will edit the `replication_info.json` to update the `scipy` version and then regenerate the Dockerfile.<<END>>
Action: edit_file: { "file_path": "./data/original/4/python/replication_info.json", "edit_type": "replace","old_text": "\"scipy==1.7.1\"", "new_text": "\"scipy==1.7.3\""}
[AGENT PROPOSES A FILE EDIT]
FULL PATH: /data/original/4/python/replication_info.json
DIFF:
--- /data/original/4/python/replication_info.json (before)
+++ /data/original/4/python/replication_info.json (after)
@@ -42,7 +42,7 @@
        "packages": {
          "python": [
            "pandas==1.3.3",
-          "scipy==1.7.1",
+          "scipy==1.7.3",
            "numpy==1.21.2"
Do you approve this edit? (yes/no): yes
Successfully edited /data/original/4/python/replication_info.json
...
<<G>>Successful Run<<END>>
\end{lstlisting}
\end{minipage}
\caption{Execution trace for resolving a Docker build failure caused by an incompatible SciPy version.}
\label{fig:scipy}
\Description[Execution trace for resolving build error.]{Execution trace for resolving a Docker build failure caused by an incompatible SciPy version.}
\end{figure}

\paragraph{Case 2: Silent Failures in Semantic Translation}
While iterative debugging fixes crashed, it cannot always ensure correct replication results.
Figure~\ref{fig:translation-error} shows a failed R-to-Python translation where the agent initially crashed due to a missing variable (\textit{`data\_clean`}).
It correctly ``fixed'' the crash by adding code to load the dataset, but it hallucinated the filename (\textit{`data\_clean.rds`} instead of \textit{`data\_clean\_5pct.rds`}).
Because the incorrect file existed in the directory, the code ran successfully, leading to a "valid" execution log but incorrect replication conclusion.


\begin{figure}[htbp!]
\centering
\begin{minipage}{0.5\textwidth}
\begin{lstlisting}[style=logstyle]
============================== DESIGN STEP =============================
---Turn 7 ---
Thought: [...] The dataset seems suitable for replication as it includes the necessary variables.
Next, I will examine the R script `data_analysis_code.R` [...]

Running Action: read_file: {"file_path": "[...]/data_analysis_code.R"}

--- Turn 8 ---
Observation: [...]
<<G>>data_clean_5pct <- readRDS("data_clean_5pct.rds")
# imputed data<<END>>
<<G>>data_imp_5pct <- readRDS("data_imp_5pct.rds")<<END>>

m1 <- lmer(trstprl_rev ~ imm_concern + happy_rev + stflife_rev + sclmeet_rev + distrust_soc +
             stfeco_rev + hincfel + stfhlth_rev + stfedu_rev +
             [...])
[...]

Thought: [...]To proceed with the replication study, I need to translate this R script into Python, ensuring compatibility with the dataset and maintaining the original analysis logic [...]

I will start by writing a new Python script file and then proceed to translate the R code into Python.
Action: write_file: {"file_path": [...]/data_analysis_code__py.py", "file_content":

import pandas as pd
import statsmodels.api as sm
import statsmodels.formula.api as smf

<<R>>AGENT FAIL TO LOAD DATASET DURING DESIGN STEP<<END>>
# Load the dataset
# Assuming the dataset is already loaded in a DataFrame named 'data_clean'

# Main analysis (complete cases, weights)
model = smf.mixedlm('trstprl_rev ~ imm_concern + happy_rev + stflife_rev + sclmeet_rev + distrust_soc +
                    stfeco_rev + hincfel + stfhlth_rev + stfedu_rev +
                   [...]])
result = model.fit()
print(result.summary())
}
[...]

============================ EXECUTION STEP ===========================
---Turn 24 ---
Observation: <<R>>line 12, in <module>\n    data=data_clean, groups=data_clean['cntry'], weights=data_clean['pspwght'])
Name
Error: name 'data_clean' is not defined\n<<END>>

Thought: [...] I need to load the dataset and assign it to `data_clean` at the beginning of the script. [...]

Action: edit_file: {
  "file_path": [...]/data_analysis_code__py.py",
  "edit_type": "insert_after",
  "anchor": "import statsmodels.formula.api as smf\n",
  "insert_text": 
  # Load the dataset
result = pyreadr.read_r('/workspace/replication_data/<<R>>data_clean.rds'<<END>>) <<R>>Load the wrong data file.<<END>>
}
[...]
<<G>>Successful Run<<END>> <<R>>But Fail to yield the correct replication results<<END>>
\end{lstlisting}
\end{minipage}
\caption{Design and execution trace of an incorrect translation attempt from R to Python.}
\label{fig:translation-error}
\Description[Design and execution trace of an incorrect translation attempt from R to Python.]{Design and execution trace of an incorrect translation attempt from R to Python.}
\end{figure}

\section{Evaluation Details}
\label{sec:app-evaluation}

\subsection{Effectiveness of LLM-as-judge}
We rely on four human evaluators to score a subset of 53 extractions from a set of six papers. 
Similarly to the LLM judge, the human evaluators are given the agent-produced post-registration and expert-annotated post-registration to assign an interger score of 0 to 3 for each extracted field  based on semantic match with the reference. 
We also provide both the human evaluators and the LLM judge with multiple examples of each score 0, 1, 2, and 3. 
We ensure that each extracted field has three annotations. 
We measure the correlation between the human evaluators and the LLM-judge, detailed in Table. \ref{tab:llmeval-effectiveness}. We also consider more cost-effective metrics like ROUGE-L \cite{lin-2004-rouge} and BERTScore \cite{zhang2019bertscore}. 
According to table~\ref{tab:llmeval-effectiveness}, LLMEval has the highest alignment with human judgement of semantic matching between the agent-produced post-registration and expert-annotated post-registration. 
Thus, we adopted LLMEval with GPT-4o as the main automated evaluation mechanism.

\begin{table}[ht!]
\centering
\begin{tabular}{lcc}
\toprule
\textbf{Method} & Spearman's correlation. & Kendall's Tau  \\
\hline
ROUGE-L & 74.66 &  60.57 \\
BERTScore & 72.38 & 57.97  \\
LLMEval (GPT-4o) & \textbf{86.18} & \textbf{78.43} \\
\bottomrule
\end{tabular}
\caption{Correlation between human assessments and automated evaluation metrics for extration stage, as indicated by Spearman $\rho$ and Kendall $\tau$ correlation coefficients. The highest scores are highlighted with bold markers.}
\label{tab:llmeval-effectiveness}
\end{table}

For design, execution, and interpretation, we collect human evaluations of 128 checkpoints across the three stages. Each checkpoint is annotated by three human evaluators with research experience, whose annotations achieve a Krippendorff's Alpha of 0.591, indicating moderate agreement and reflecting the complexity of evaluating scientific replications. We compare these human judgments with GPT-4o, and GPT-5 judgments to test whether the choice of LLM affects the resulting scores and whether GPT-4o received an inflated score when judged by itself. GPT-4o aligns closer with human judgment than GPT-5, with a Pearson correlation coefficient of 0.659 compared to 0.525 for GPT-5. The correlation between GPT-4o and GPT-5 judgments is lower at 0.403, suggesting that the choice of LLM can meaningfully affect automated evaluation. We therefore use GPT-4o as the main LLM judge and include GPT-5 scores only as robustness check. 

\subsection{Evaluation prompts and templates}


For evaluation of the extraction, generation-design, and interpretation stages, we perform a single call to the LLM. 
For the generation-execution stage, the evaluation criteria require an inspection of the execution logs, so we adopt a ReAct \cite{yao2022react} implementation for the GPT-4o judge. 
The full evaluation prompts, checkpoints, and templates are provided in our supplementary artifact~\cite{artifact}.







\subsection{Additional Results}
\label{app:additional-results}

\begin{table}[ht!]
\centering
\begin{tabular}{l|c|c|c|c}
\toprule
\textbf{Model} & \textbf{Extract - Info} & \textbf{Design} & \textbf{Execute} & \textbf{Interpret}  \\
\midrule
\lstinline{o3}         & \textbf{60.25}  & \textbf{54.16} & 88.35 & 72.11  \\
\lstinline{GPT-4o}    & 55.93 & 48.79 & 71.47 & 53.40 \\
\lstinline{GPT-5-mini} & 59.65 & 42.65 & 64.43 & 73.58  \\
\lstinline{GPT-5}     & 58.91 & 53.95 & \textbf{94.26}   & \textbf{77.42}  \\
\bottomrule
\end{tabular}%
\caption{Performance of our LLM agents across different stages against {\sc \textbf{ReplicatorBench}} in Python mode. GPT-5 is used as the backbone model for LLMEval. All evaluation scores are normalized to a 0-100 scale. Macro aggregations are used. Results for \textit{Extract - Web} and \textit{Outcome}  are deterministic and therefore remain the same as those reported in Table~\ref{tab:merged_results}.} 
\label{tab:merged_results-gpt-5}
\end{table}

\begin{table}[ht!]
\centering
\resizebox{\linewidth}{!}{%
\begin{tabular}{lc|c|c|c|c}
\toprule
\textbf{Model} & \textbf{Code Mode} & \textbf{Design} & \textbf{Execute} & \textbf{Interpret} & \textbf{Outcome}  \\
\midrule
\multirow{2}{*}{\lstinline{o3}} & native       & 78.61	& 69.01 &	87.82 &	55.56  \\
 & python       & \textbf{84.57}	& \textbf{95.10}	& \textbf{95.48} & 	\textbf{62.35}  \\
\multirow{2}{*}{\lstinline{GPT-4o}} & native  & 78.47 &	60.33	& 76.31	& 45.45 \\
 & python   & \textbf{81.14} &	\textbf{74.10}	& \textbf{77.01}	& \textbf{59.82}  \\
\multirow{2}{*}{\lstinline{GPT-5}} & native    & 81.36	& 77.96 &	\textbf{91.18} &	74.54  \\
 & python    & \textbf{83.10}	& \textbf{94.44} &	77.73 &	\textbf{84.17} \\
\bottomrule
\end{tabular}%
}
\caption{Performance of {\sc \textbf{ReplicatorAgent}} in a Python-only setting compared to Native setting. LLMEval scores are reported for design, execution, and interpretation stage. Macro F1 Scores are reported for the final replication outcome (criteria met or unmet). The best results for each model are highlighted in bold markers.}
\label{tab:native-vs-python}
\end{table}

\section{Task Templates, Agent Prompts, and Artifacts}
\label{sec:task-templates}
To support reproducibility while respecting the page limit, we provide all prompts and structured output schemas in a supplementary artifact.\footnote{\url{https://doi.org/10.5281/zenodo.20506946}}
It contains the core ReAct agent system prompt, each specific prompts for each stage and structured output schemas for each stage in our pipeline.

We provide an example for the \emph{alias matching} technique in the Extraction stage. If the ground truth URL is
\begin{quote}
    \href{https://www2.census.gov/programs-surveys/popest/datasets/2010-2019/counties/totals/}{https://www2.census.gov/programs-surveys/popest/datasets/2010-2019/counties/totals/},
\end{quote}
then we accept
\begin{quote}
\href{https://www2.census.gov/}{https://www2.census.gov/}.
\end{quote}

\end{document}